\documentclass[11pt,a4paper]{article}
\usepackage[hyperref]{emnlp-ijcnlp-2019}
\usepackage{times}
\usepackage{latexsym}
\usepackage[normalem]{ulem}

\aclfinalcopy 

\usepackage{multirow}
\usepackage{arydshln}
\usepackage{amsmath}

\usepackage{algorithm}
\usepackage{algorithmic}

\usepackage{subfigure}

\usepackage{booktabs}

\usepackage{url}

\usepackage{graphicx}

\newcommand{\model}{DAFE}

\title{Unsupervised Domain Adaptation for Neural Machine Translation with Domain-Aware Feature Embeddings}

\author{Zi-Yi Dou, Junjie Hu, Antonios Anastasopoulos, Graham Neubig \\
  Language Technologies Institute, Carnegie Mellon University \\
  {\tt \{zdou, junjieh, aanastas, gneubig\}@cs.cmu.edu}}
\date{}

\begin{document}
\maketitle
\begin{abstract}
  The recent success of neural machine translation models relies on the availability of high quality, in-domain data. Domain adaptation is required when domain-specific data is scarce or nonexistent. Previous unsupervised domain adaptation strategies include training the model with in-domain copied monolingual or back-translated data. However, these methods use generic representations for text regardless of domain shift, which makes it infeasible for translation models to control outputs conditional on a specific domain. In this work, we propose an approach that adapts models with domain-aware feature embeddings, which are learned via an auxiliary language modeling task. Our approach allows the model to assign domain-specific representations to words and output sentences in the desired domain. Our empirical results demonstrate the effectiveness of the proposed strategy, achieving consistent improvements in multiple experimental settings. In addition, we show that combining our method with back translation can further improve the performance of the model.\footnote{Our code is publicly available at: \url{https://github.com/zdou0830/DAFE.}}

　
\end{abstract}


\section{Introduction}
\label{sec:introduction}


While neural machine translation (NMT) systems have proven to be effective in scenarios where large amounts of in-domain data are available~\cite{Gehring:2017:ICML,Vaswani:2017:NIPS,chen2018best}, they have been demonstrated to perform poorly when the test domain does not match the training data~\cite{koehn2017six}. 
Collecting large amounts of parallel data in all possible domains we are interested in is costly, and in many cases impossible.
Therefore, it is essential to explore effective methods to train models that generalize well to new domains. 

\begin{figure}[t]
\centering
\includegraphics[width=0.49\textwidth]{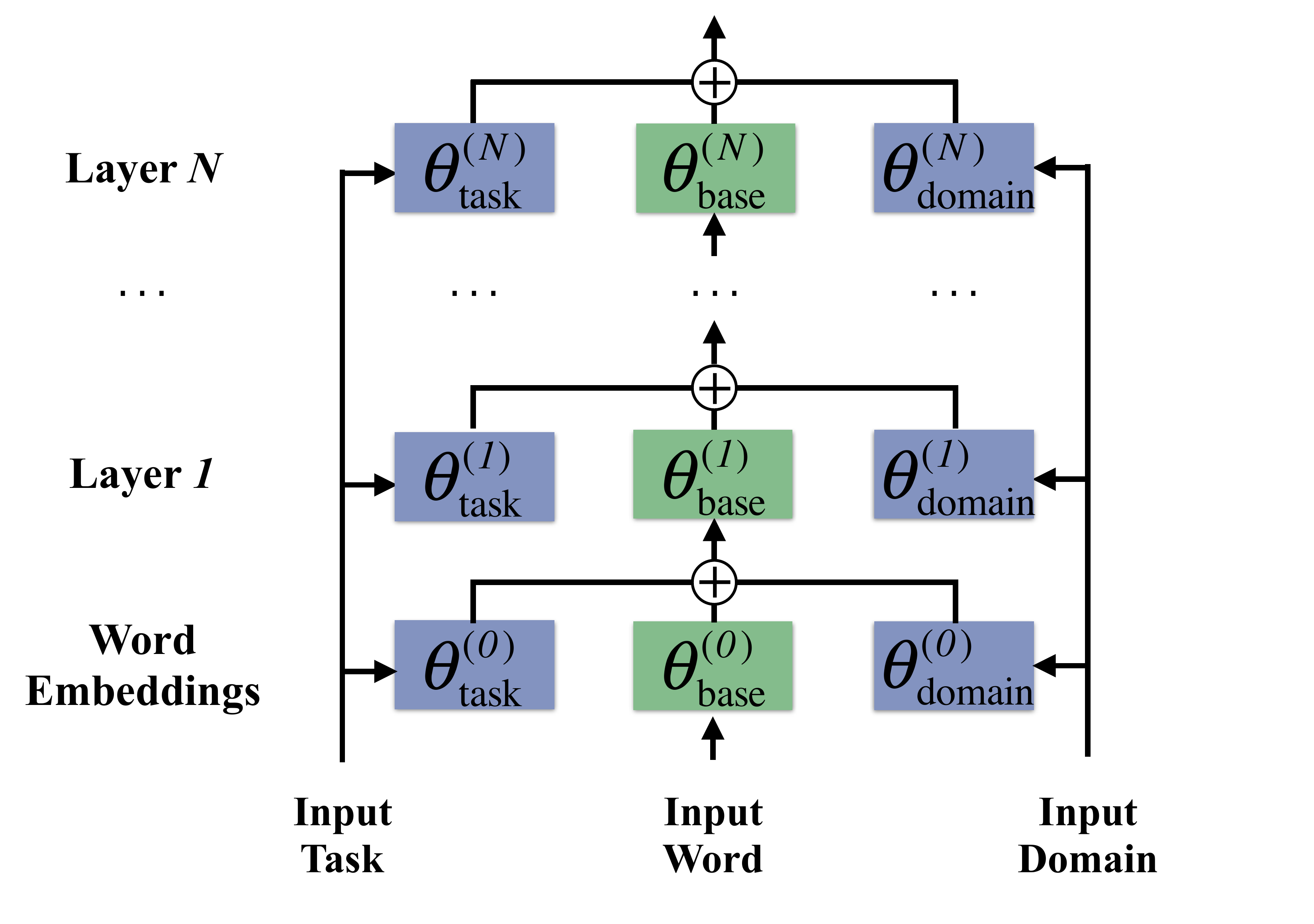}
\caption{Main architecture of {\model}. Embedding learners generate domain- and task-specific features at each layer, which are then integrated into the output of the base network.}
\label{fig:arch}
\end{figure}

Domain adaptation for neural machine translation has attracted much attention in the research community, with the majority of work focusing on the supervised setting where a small amount of in-domain data is available~\cite{luong2015stanford,freitag2016fast,chu2017empirical,vilar2018learning}.
An established approach is to use domain tags as additional input, with the domain representations learned over parallel data \cite{kobus2017domain}.
In this work, we focus on \emph{unsupervised} adaptation, where there are no in-domain parallel data available. 
Within this paradigm, \newcite{currey2017copied} copy the in-domain monolingual data from the target side to the source side and \newcite{sennrich2016improving} concatenate back-translated data with the original corpus. 
{However, these methods learn generic representations for all the text, as the learned representations are shared for all the domains and synthetic and natural data are treated equally. Sharing embeddings may be sub-optimal as data from different domains are inherently different. This problem is exacerbated when words have different senses in different domains. }

{In this work, we propose a method of {\it Domain-Aware Feature Embedding} {(\model)} that performs unsupervised domain adaptation by disentangling representations into different parts. 
Because we have no in-domain parallel data, we learn {\model}s via an auxiliary task, namely language modeling. Specifically, our proposed model consists of a base network, whose parameters are shared across settings, 
as well as both domain and task embedding learners. By separating the model into different components, {\model} can learn representations tailored to specific domains and tasks, which can then be utilized for domain adaptation.} 


We evaluate our method in a Transformer-based NMT system~\cite{Vaswani:2017:NIPS} under two different data settings. Our approach demonstrates consistent improvements of up to 5 BLEU points over unadapted baselines, and up to 2 BLEU points over strong back-translation models. Combining our method with back translation can further improve the performance of the model, suggesting the orthogonality of the proposed approach and methods that rely on synthesized data. 

\vspace{-2pt}
\begin{table*}[ht]
  \centering
  \resizebox{\textwidth}{!}{
  \begin{tabular}{ll||c|c|c|c|c|c||c|c|c|c|c|c}
   \multicolumn{2}{c||}{\multirow{3}{*}{\bf Method} } & \multicolumn{6}{c||}{\bf German-English} &  \multicolumn{3}{c|}{\bf Czech-English} & \multicolumn{3}{c}{\bf German-English}  \\ 
    \cline{3-14}
  & & \multicolumn{2}{c|}{LAW} & \multicolumn{2}{c|}{MED} &  \multicolumn{2}{c||}{IT}   & \multicolumn{3}{c|}{WMT} & \multicolumn{3}{c}{WMT}  \\
   \cline{3-14}
  & & MED & IT & LAW & IT & LAW & MED  & TED & LAW & MED  & TED & LAW & MED\\
  \hline
  \hline
  (1)&Baseline & 13.25 & 5.22 & 4.55 & 3.15 & 4.29 & 7.05 & 24.30 & 28.22 & 15.45 & 28.15 & 24.61 & 26.75 \\
    \hline
    \multicolumn{14}{l}{\textit{Data-Centric Methods}} \\
    \hline
   (2)&Copy & 19.23 & 7.57 & 6.01 & 5.89 & 5.11 & 11.15 & 26.44 & 36.49 & 22.73 & 29.79 & 26.17 & 29.33\\
   (3)&Back & 22.53 & 11.34 & 7.62 & 7.02 & 8.06 & 14.56 & 32.70 & 42.08 & 30.45 & 34.46 & 30.24 & 33.16 \\
   \hline
    \multicolumn{14}{l}{\textit{Model-Centric Methods}} \\
    \hline
   (4)&{\model} w/o Embed & 23.44 & 6.97 & 9.21 & 8.34 & 8.09 & 16.34 & 26.63 & 35.86 & 23.44 & 29.88 & 27.10 & 32.15 \\
   (5)&{\model} & 24.23 & 8.59 & 9.87 & 8.44 & 8.61 & 17.50 & 28.09 & 38.89 & 26.05 & 30.88 & 27.77 & 32.48 \\
   \hline
    \multicolumn{14}{l}{\textit{Combining Data-Centric and Model-Centric Methods}} \\
    \hline
   (6)&Back + {\model} & 25.34 & 13.55 & 9.60 & 11.20 & 9.60 & 17.25 & \bf 33.18 & 44.06 & 34.24 & 34.57 & 30.72 & 35.48 \\
   (7)&Back-{\model} & 26.47 & 13.75 & 11.90 & 9.47 & 10.60 & 18.04 & 32.51 & 43.33 & 35.45 & 34.57 & 30.93 & 37.66\\
   (8)&\multicolumn{1}{l||}{Back-{\model} + {\model}} & \bf 26.96 & \bf 15.41 & \bf 14.28 & \bf 13.03 & \bf  11.67 &  \bf 21.30 & 33.02 & \bf 44.36 & \bf 37.48 & \bf 34.89 & \bf  31.46 & \bf  38.79\\
    \end{tabular}}
    \caption{\label{tab:main} Translation accuracy (BLEU) under different settings. 
    The second and third rows list source and target domains respectively.  ``{\model} w/o Embed'' denotes {\model} without embedding learners and ``Back-{\model}'' denotes back-translation by target-to-source model trained with {\model}. {\model} outperforms other approaches when adapting between domains (row~1-5, column~2-7) and is complementary to back-translation (row~6-8).
    }
  \end{table*}

\section{Methods}\label{sec:methods}

In this section, we first illustrate the architecture of {\model}, then describe the overall training strategy.

\subsection{Architecture}
{\model} disentangles hidden states into different parts so that the network can learn representations for particular domains or tasks, as illustrated in Figure~\ref{fig:arch}.
Specifically, it consists of three parts: a {\it base network} with parameters $\theta_{\text{base}}$ that learns common features across different tasks and domains, a {\it domain-aware feature embedding learner} 
that generates embeddings $\theta^{\tau}_{\text{domain}}$ given input domain $\tau$ 
 and a {\it task-aware feature embedding learner} 
that outputs task representations $\theta^{\gamma}_{\text{task}}$ given input task $\gamma$.  The final outputs for each layer are obtained by a combination of the base network outputs and feature embeddings.

 The base network is implemented in the encoder-decoder framework~\cite{seq2seq,cho2014learning}. Both the task and domain embedding learners directly output feature embeddings with look-up operations. 

In this work, the domain-aware embedding learner learns domain representations $\theta^{in}_{\text{domain}}$ and $\theta^{out}_{\text{domain}}$ from in-domain and out-of-domain data respectively, and the task-aware embedding leaner learns task embeddings  $\theta^{mt}_{\text{task}}$ and $\theta^{lm}_{\text{task}}$ for machine translation and language modeling.


The feature embeddings are generated at \emph{each} encoding layer (including the source word embedding layer) and have the same size as the hidden states of the base model. It should be noted that feature embedding learners generate different embeddings at different layers. 

Formally, given a specific domain $\tau$ and task $\gamma$, the output of the $l$-th encoding layer ${\bf H}_e^{(l)}$ would be:
\begin{equation*}
    {\bf H}_e^{(l)} = \textsc{Layer}_e({\bf H}_e^{(l-1)}; \theta_{\text{base}}^{(l)}) + {\theta}_{\text{domain}}^{\gamma, (l)} + {\theta}_{\text{task}}^{\tau, (l)},
\end{equation*}
where ${\theta}_{\text{domain}}^{\gamma,(l)}$ and ${\theta}_{\text{task}}^{\tau, (l)}$ are single vectors and $\textsc{Layer}_e(\cdot)$ can be any layer encoding function, such as an LSTM~\cite{hochreiter1997long} or Transformer~\cite{Vaswani:2017:NIPS}. 

In this paper, we adopt a simple, add operation to combine outputs of different parts which already achieves satisfactory performance as shown in Section~\ref{sec:experiment}. We leave investigating more sophisticated combination strategies for future work.

\begin{algorithm}[t]
\caption{\label{alg:main}Training Strategy}
\begin{algorithmic}[1]
    \WHILE{$\theta_{\text{base}},\theta_{\text{domain}},\theta_{\text{task}}$ have not converged}
		\STATE{Sample $\{(C(\mathbf{y}), \mathbf{y})\}$ from $Y^{in} $} 
		\STATE{Train $\{\theta_{\text{base}},\theta_{\text{domain}}^{in}, \theta_{\text{task}}^{lm}\}$ with Eqn. \ref{eqn:lm}}	
		\STATE{Sample $\{(C(\mathbf{y}), \mathbf{y})\}$ from $Y^{out}$} 
		\STATE{Train $\{\theta_{\text{base}},\theta_{\text{domain}}^{out}, \theta_{\text{task}}^{lm}\}$ with Eqn. \ref{eqn:lm}}
		\STATE{Sample $\{(\mathbf{x}, \mathbf{y})\}$ from $(X^{out}, Y^{out}) $}
		\STATE{Train $\{\theta_{\text{base}},\theta_{\text{domain}}^{out}, \theta_{\text{task}}^{mt}\}$ with Eqn. \ref{eqn:mt}}
	\ENDWHILE
\end{algorithmic}
\end{algorithm}

\subsection{Training Objectives}

In the unsupervised domain adaptation setting, we assume access to an out-of-domain parallel training corpus $(X^{out}, Y^{out})$ and target-language in-domain monolingual data $Y^{in}$.

\paragraph{Neural machine translation.}
Our target task is machine translation. 
Both the base network and embedding learners are jointly trained with the objective:
\begin{equation}
\label{eqn:mt}
    \max_{\theta } \sum_{(\mathbf{x}, \mathbf{y}) \in (X^{out}, Y^{out}) } \log p(\mathbf{y} | \mathbf{x}; \theta ),
\end{equation}
where $\theta = \{ \theta_{\text{base}},\theta_{\text{task}}^{mt},\theta_{\text{domain}}^{out} \}$.

\paragraph{Language modeling.}
We choose masked language modeling (LM) as our auxiliary task. Following ~\newcite{lample2018unsupervised,lample2018phrase}, we create corrupted versions $C(\mathbf{y})$ for each target sentence $\mathbf{y}$ by randomly dropping and slightly shuffling words. During training, gradient ascent is used to maximize the objective:
\begin{equation}
\label{eqn:lm}
    \max_{\theta} \sum_{\mathbf{y} \in \{ Y^{in} \cup Y^{out} \} } \log p(\mathbf{y} | C(\mathbf{y}) ; \theta),
\end{equation}
where $\theta = \{ \theta_{\text{base}},\theta_{\text{task}}^{lm},\theta_{\text{domain}}^{out} \}$ for out-of-domain data and $\{ \theta_{\text{base}},\theta_{\text{task}}^{lm},\theta_{\text{domain}}^{in} \}$ for in-domain data. 

\paragraph{Training strategy.}

Our training strategy is shown in Algorithm~\ref{alg:main}. The ultimate goal is to learn a set of parameters $\{\theta_{\text{base}}, \theta_{\text{domain}}^{in}, \theta_{\text{task}}^{mt}\}$ for in-domain machine translation. While out-of-domain parallel data allows us to train 
 $\{\theta_{\text{base}}, \theta_{\text{domain}}^{out}, \theta_{\text{task}}^{mt}\}$, the monolingual data help the model learn both $\theta_{\text{domain}}^{in}$ and $\theta_{\text{domain}}^{out}$.


\section{Experiments}
\label{sec:experiment}

  \vspace{-2pt}
\subsection{Setup}

  \vspace{-2pt}

\paragraph{Datasets.}
We validate our models in two different data settings. First, we train on the law, medical and IT datasets of the German-English OPUS corpus~\cite{tiedemann2012parallel} and test our methods' ability to adapt from one domain to another. The dataset contain 2K development and test sentences in each domain, and about 715K, 1M and 337K training sentences respectively. These datasets are relatively small and the domains are quite distant from each other. In the second setting, we adapt models trained on the general-domain WMT-14 datasets into both the TED~\cite{duh18multitarget} and law, medical OPUS datasets. For this setting, we consider two language pairs, namely Czech and German to English. The Czech-English and German-English datasets consist of 1M and 4.5M sentences and the development and test sets contain about 2K sentences.

\paragraph{Models.} We implement {\model} on top of the Transformer model. Both the encoder and decoder consist of 4 layers and the hidden size is set to 512. 
Byte-pair encoding~\cite{sennrich2016neural} is employed to process training data into subwords for a final shared vocabulary size of 50K.

\paragraph{Baselines.}
We compare our methods with two baseline models: 1) The copied monolingual data model~\cite{currey2017copied} which copies target in-domain monolingual data to the source side; 2) Back-translation~\cite{sennrich2016improving} which enriches the training data by generating synthetic in-domain parallel data via a target-to-source NMT model. We characterize the two baselines as {\emph{data-centric}} methods as they rely on synthesized data. In contrast, our approach is \emph{model-centric} as we mainly focus on modifying the model architecture. We also perform an ablation study by removing the embedding learners (denoted as ``DAFE w/o Embed'') and the model will just perform multi-task learning.
  \vspace{-2pt}
\subsection{Main Results}

  \vspace{-2pt}
\paragraph{Adapting between domains.}
  As shown in the first 6 results columns of Table~\ref{tab:main}, the unadapted baseline model (row 1) performs poorly when adapting between domains. The copy method (row~2) and back translation (row~3) both improve the model significantly, with back-translation being a better alternative to copying. {\model} (row~5) achieves superior performance compared to back-translation, with improvements of up to~2 BLEU points. Also, removing the embedding learners leads to degraded performance (row~4), indicating the necessity of their existence. 
  
  \vspace{-2pt}
  
  \paragraph{Adapting from a general to a specific domain.} In the second data setting (last~6 columns of Table~\ref{tab:main}), with relatively large amounts of general-domain datasets, back-translation achieves competitive performance. In this setting, {\model} improves the unadapted baseline significantly but it does not outperform back-translation. We hypothesize this is because the quality of back-translated data is relatively good.

\begin{figure}[t]
\centering
\includegraphics[width=.8\columnwidth]{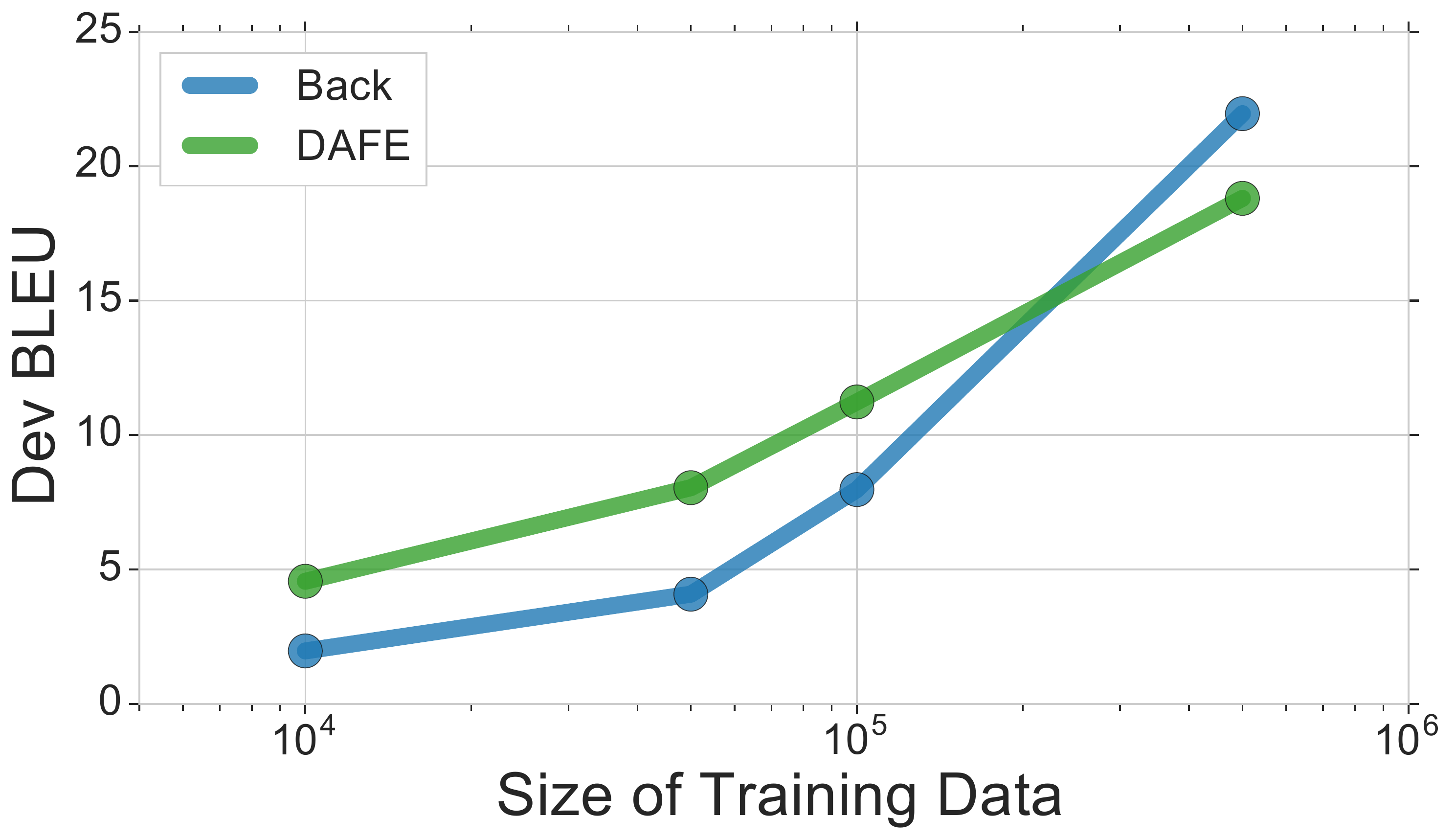}
\caption{{\model} outperforms back-translation in low-resource scenarios.} 
\label{fig:back}
\end{figure}

\subsection{Combining {\model} with Back-Translation}
We conjecture that {\model} is complementary to the data-centric methods.
We attempt to support this intuition by combining \model{} with back-translation (the best data-centric approach).
We try three different strategies to combine {\model} with back-translation, outlined in Table~\ref{tab:main}. 
  
Simply using back-translated data to train {\model} (row~6) already achieves notable improvements of up to 4 BLEU points. 
We can also generate back-translated data using target-to-source models trained with {\model}, with which we train the forward model (Back-{\model}, row~7). By doing so, the back-translated data will be of higher quality and thus the performance of the source-to-target model can be improved.
The overall best strategy is to use Back-{\model} to generate synthetic in-domain data and train the {\model} model with the back-translated data (Back{-\model}+{\model}, row~8).
Across almost all adaptation settings, Back{-\model}+{\model} leads to higher translation quality, as per our intuition.
An advantage of this setting is that the back-translated data allow us to learn $\theta_{\text{domain}}^{in}$ with the translation task.

\subsection{Analysis} 
\paragraph{Low-resource scenarios.}
One advantage of {\model} over back-translation is that we do not need a good target-to-source translation model, which can be difficult be acquire in low-resource scenarios. We randomly sample different amounts of training data and evaluate the performance of {\model} and back-translation on the development set. As shown in Figure~\ref{fig:back}, {\model} significantly outperforms back-translation in data-scarce scenarios, as low quality back-translated data can actually be harmful to downstream performance.
   
  \paragraph{Controlling the output domain.}
  
  \begin{table}[t]
  \centering
  \resizebox{0.45\textwidth}{!}{%
  \begin{tabular}{c|cc|cc}
   \bf Embedding & MED dev & MED test & IT dev & IT test \\
   \hline
   \hline
   MED & 42.06 & 34.63 & 4.13 & 4.80\\
   \hdashline
   IT  & 36.96 & 30.09 & 7.54 & 8.44\\
    \end{tabular}}
    \caption{\label{tab:control} Providing mismatched domain embeddings leads to degraded performance. }
  \end{table}

  An added perk of our model is the ability to control the output domain by providing the desired domain embeddings. 
  As shown in Table~\ref{tab:control}, feeding mismatched domain embeddings leads to worse performance. Examples in Table~\ref{tab:examples} further suggest the model with medical embeddings as input can generate domain-specific words like ``EMEA'' (European Medicines Evaluation Agency) and ``intramuscular'', while IT embeddings encourage the model to generate words like ``bug'' and ``developers''.
  
   \begin{table}[t]
  \centering
    \resizebox{0.5\textwidth}{!}{
  \begin{tabular}{|l|l|}
  \hline
 \bf Reference & please report this bug to the developers . \\
 \hline
 {\bf MED-embed} & please report this to the EMEA .  \\
 \hdashline
\bf IT-embed & please report this bug to the developers .\\
    \hline
    \hline
    \bf Reference & for intramuscular use . \\
 \hline
 {\bf MED-embed} & {for intramuscular use .} \\
 \hdashline
\bf IT-embed & for the use of the product .\\
\hline
    \end{tabular}
    }
    \caption{\label{tab:examples} Controlling the output domain by providing different domain embeddings. We use compare-mt~\cite{neubig19naacl} to select examples.}
  \end{table}

  
\vspace{-2pt}
\section{Related Work}
\label{sec:related}
\vspace{-2pt}
Most previous domain adaptation work for NMT focus on the setting where a small amount of in-domain data is available. Continued training~\cite{luong2015stanford,freitag2016fast} methods first train an NMT model on out-of-domain data and then fine-tune it on the in-domain data. Similar to our work,~\newcite{kobus2017domain} propose to use domain tags to control the output domain, but it still needs a in-domain parallel corpus and our architecture allows more flexible modifications than just adding additional tags.

Unsupervised domain adaptation techniques for NMT can be divided into data- and model-centric methods~\cite{chu2018survey}. 
Data-centric approaches mainly focus on selecting or generating the domain-related data using existing in-domain monolingual data. 
Both the copy method
~\cite{currey2017copied} and back-translation~\cite{sennrich2016improving} are representative data-centric methods. In addition, \newcite{moore2010intelligentselection,axelrod2011adaptation,duh2013adaptation} use LMs to score the out-of-domain data, based on which they select data similar to in-domain text.
Model-centric methods have not been fully investigated yet. \newcite{gulcehre2015using} propose to fuse LMs and NMT models, but their methods require querying two models during inference and have been demonstrated to underperform the data-centric ones~\cite{chu2018comprehensive}.
There are also work on adaptation via retrieving sentences or n-grams in the training data similar to the test set~\cite{farajian2017multi,bapna2019non}. However, it can be difficult to find similar parallel sentences in domain adaptation settings. 

 \vspace{-2pt}
\section{Conclusion}
\label{sec:conclusion}
\vspace{-2pt}
In this work, we propose a simple yet effective unsupervised domain adaptation technique for neural machine translation, which adapts the model by domain-aware feature embeddings learned with language modeling. Experimental results demonstrate the effectiveness of the proposed approach across settings. In addition, analysis reveals that our method allows us to control the output domain of translation results. Future work include designing more sophisticated architectures and combination strategies as well as validating our model on other language pairs and datasets. 

\section*{Acknowledgements}
We are grateful to Xinyi Wang and anonymous reviewers for their helpful suggestions and insightful comments. We also thank Zhi-Hao Zhou, Shuyan Zhou and Anna Belova for proofreading the paper.

This material is based upon work generously supported partly by the National Science Foundation under grant 1761548 and the Defense Advanced Research Projects Agency Information Innovation Office (I2O) Low Resource Languages for Emergent Incidents (LORELEI) program under Contract No. HR0011-15-C0114. The views and conclusions contained in this document are those of the authors and should not be interpreted as representing the official policies, either expressed or implied, of the U.S. Government. The U.S. Government is authorized to reproduce and distribute reprints for Government purposes notwithstanding any copyright notation here on.


\bibliography{emnlp-ijcnlp-2019}
\bibliographystyle{acl_natbib}

\cleardoublepage

\end{document}